# UnPaSt: unsupervised patient stratification by biclustering of omics data

Michael Hartung[1], Andreas Maier[1], Yuliya Burankova[1,2], Fernando Delgado-Chaves[1], Olga I. Isaeva[3], Alexey Savchik[4], Fábio Malta de Sá Patroni[5], Jens J. G. Lohmann[1], Daniel He[6], Casey Shannon[6], Jan-Ole Schulze[1], Katharina Kaufmann[1], Zoe Chervontseva[1], Farzaneh Firoozbakht[1], Anne Hartebrodt[7], Niklas Probul[1], Olga Tsoy[1,8], Alexandra Abisheva[9], Evgenia Zotova[9], Kavya Singh[10], Kristel Van Steen[10], Malte Kuehl[11,12], Victor G. Puelles[11,12,13,14], David B. Blumenthal[7], Martin Ester[15,16], Tanja Laske[1,17], Jan Baumbach[1,18], Olga Zolotareva[1,19].

[1] *Institute for Computational Systems Biomedicine, University of Hamburg, Albert-Einstein-Ring 10, Hamburg, Germany*
[2] *Chair of Proteomics and Bioanalytics, TUM School of Life Sciences, Technical University of Munich, Freising, Germany*
[3] *Division of Tumor Biology & Immunology, Netherlands Cancer Institute, Plesmanlaan 121, Amsterdam, the Netherlands*
[4] *ACMetric, Amsterdam, the Netherlands.*
[5] *University of Campinas, Campinas, Brazil.*
[6] *University of British Columbia, Vancouver, Canada.*
[7] *Biomedical Network Science Lab, Department Artificial Intelligence in Biomedical Engineering, Friedrich-Alexander-Universität Erlangen-Nürnberg, Erlangen, Germany*
[8] *Computer Science Department, Faculty of Science, Vrije Universiteit Amsterdam, De Boelelaan 1111, 1081 HV Amsterdam*
[9] *Altius Institute for Biomedical Sciences, Seattle, United States.*
[10] *GIGA - Molecular and Computational Biology, University of Liège, Liège, Belgium.*
[11] *Department of Clinical Medicine, Aarhus University, 8200 Aarhus, Denmark.*
[12] *Department of Pathology, Aarhus University Hospital, 8200 Aarhus N, Denmark.*
[13] *III. Department of Medicine, University Medical Center Hamburg-Eppendorf, Hamburg, Germany*
[14] *Hamburg Center for Kidney Health (HCKH), Hamburg, Germany*
[15] *Simon Fraser University, Burnaby, Canada*
[16] *Vancouver Prostate Centre, Vancouver, Canada*
[17] *Viral Systems Modeling, Leibniz Institute of Virology, 20251 Hamburg, Germany*
[18] *Department of Mathematics and Computer Science, University of Southern Denmark, Odense, Denmark.*
[19] *Data Science in Systems Biology, TUM School of Life Sciences, Technical University of Munich, Freising, Germany*

# Abstract

Unsupervised patient stratification is essential for disease subtype discovery, yet, despite growing evidence of molecular heterogeneity of non-oncological diseases, popular methods are benchmarked primarily using cancers with mutually exclusive molecular subtypes well-differentiated by numerous biomarkers.

Evaluating 22 unsupervised methods, including clustering and biclustering, using simulated and real transcriptomics data revealed their inefficiency in scenarios with non-mutually exclusive subtypes or subtypes discriminated only by few biomarkers.

To address these limitations and advance precision medicine, we developed UnPaSt, a novel biclustering algorithm for unsupervised patient stratification based on differentially expressed biclusters. UnPaSt outperformed widely used patient stratification approaches in the *de novo* identification of known subtypes of breast cancer and asthma. In addition, it detected many biologically insightful patterns across bulk transcriptomics, proteomics, single-cell, spatial transcriptomics, and multi-omics datasets, enabling a more nuanced and interpretable view of high-throughput data heterogeneity than traditionally used methods.

# Introduction

Historically, classifications of human diseases are based on symptoms and affected organs. However, large-scale clinical and omics studies have revealed that many diseases represent groups of disorders with similar manifestations but distinct molecular mechanisms[1,2]. Molecular heterogeneity is well known for cancers, many of which are now subdivided into clinically relevant subtypes based on omics data[3]. Increasing evidence now indicates that numerous non-oncological diseases also consist of molecularly distinct subtypes[4–8]. This hidden disease heterogeneity likely contributes to inter-individual variability of drug response[9], low success rate in translating preclinical results into clinical applications[10], and confounds downstream analyses, thereby obscuring the elucidation of disease mechanisms. Because the number of molecular subtypes and their defining features are typically unknown *a priori*, their discovery is commonly performed using unsupervised machine learning methods, enabling an unbiased exploration of data heterogeneity.

Most commonly used strategies, such as k-means, hierarchical clustering, or matrix factorization methods[11–13] have been particularly successful in transcriptome-based stratification of cancers, where subtypes are well-differentiated by a large number of differentially expressed genes[14,15]. However, these methods explore global similarity of samples based on all features and perform poorly when subtype-specific biomarkers are few or when subtypes are not mutually exclusive and individual samples may belong to multiple subgroups, as is often the case in high-dimensional biological data.

Biclustering offers a promising alternative by searching for subsets of features that form local patterns independently distinguishing sample subgroups[16–20]. Unlike most clustering methods, which divide all samples into $k$ disjoint subsets, biclustering can still be effective in scenarios with a few subtype-specific biomarkers or with overlapping patterns (Figure 1). Yet, despite their conceptual benefits, biclustering methods have been largely overlooked in recent patient stratification benchmarks, which mainly focus on multi-omics data integration methods[21–26] and on cancers with pronounced differences between subtypes[22–26]. A likely reason is that most existing biclustering methods were designed to detect patterns of differential co-expression, whereas disease subtypes are often defined by differential abundance of specific biomarkers[4–8].

To overcome this limitation, we introduce UnPaSt, a novel biclustering method for <u>un</u>supervised <u>pa</u>tient <u>st</u>ratification, targeted specifically at differentially expressed biclusters. We benchmarked UnPaSt against 22 unsupervised learning methods, including both popular clustering and data integration techniques[27–39] and biclustering methods[40–45] selected for their popularity and availability of executable command line implementations (Supplementary Table S1), using simulated and real bulk transcriptomics datasets. Besides the breast cancer stratification problem addressed in many related works[21,22,24–26], we evaluated the methods for the stratification of patients with asthma, a common disease in which molecular subtypes are defined by only a few biomarkers[4]. Unlike previous studies relying on proxy performance measures, such as associations with clinical features[24,26], we directly evaluated the capabilities to *de novo* discover established molecular subtypes and evaluated the effect of parameter tuning.

In the breast cancer and asthma datasets, UnPaSt not only identified known molecular subtypes more accurately than other tested methods, but also detected several other biologically relevant expression signatures, replicable in independent datasets. We further demonstrated the utility of UnPaSt by applying it to a wide range of datasets, including multiple cancer transcriptomes, longitudinal sequencing data of deletion-containing viral genomes, proteomics, multi-omics, single-cell, and spatial transcriptomics data. Across these applications, UnPaSt consistently revealed biologically meaningful and reproducible patterns, highlighting its broad utility for exploring heterogeneity in high-dimensional data for patient stratification and other tasks.

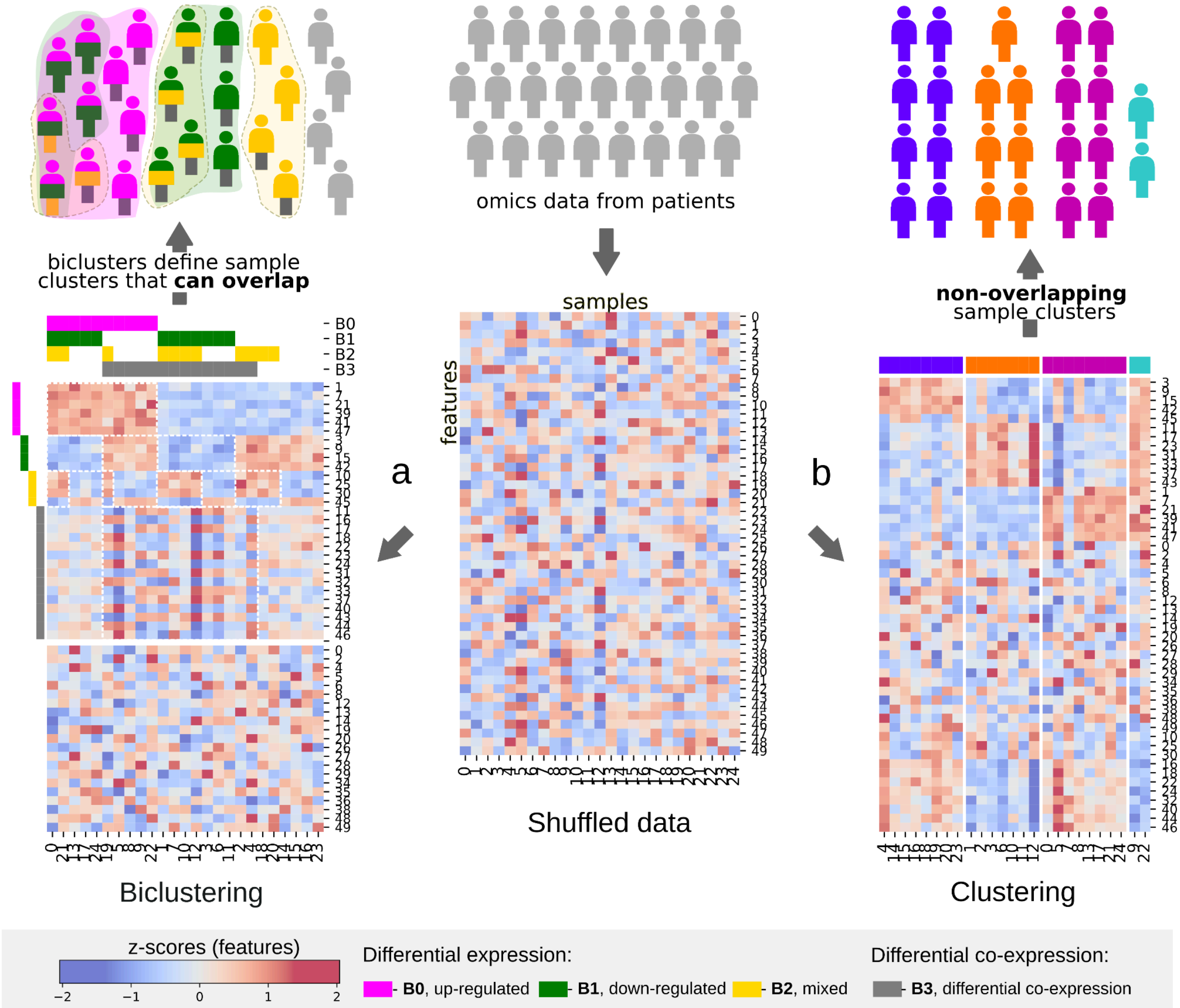

**Figure 1.** Biclustering (**a**, left) and clustering (**b**, right) of a toy data matrix with 50 rows (features) and 25 columns (samples) and including four implanted biclusters. Color bars on the heatmaps' margins show element membership in the biclusters (left) and clusters (right). Additionally, biclusters are highlighted with white dashed frames. The first three biclusters (B0, B1, and B2) correspond to the differential expression patterns, each independently dividing all columns (samples) into two subgroups with distinct feature values. The fourth bicluster B3 corresponds to a differential co-expression pattern. Rows belonging to this bicluster demonstrate strong pairwise correlation across bicluster columns but weak correlation across the columns outside it.

# Results

## UnPaSt Algorithm

Given a data matrix of $S$ samples (columns) and $F$ features (rows, e.g., gene expressions), UnPaSt identifies subsets of features $F_1, \ldots, F_i, \ldots$, splitting $S$ into two well-separated subsets $S_{F_i}$ and $\overline{S_{F_i}} = S \setminus S_{F_i}$ with distinct values for each $j$-th feature $f_j \in F_i$. For simplicity, we will further refer to the submatrix including a smaller sample set $S_{F_i}$ as "bicluster" $B_i = (F_i, S_{F_i})$ and to its complement as the "background" (Figure 2a).

Such biclusters in transcriptome data may correspond to sample sets $S_{F_i}$ that over- or under-express a certain pathway $F_i$. Because bicluster and background sample sets should be identified jointly with the feature subsets that differentiate them, this problem is related to the unsupervised formulation of the differential expression analysis task. To highlight this, we will further refer to biclusters identified by UnPaSt as *differentially expressed biclusters*. Of note, UnPaSt does not guarantee that each feature $f_j \in F_i$ is statistically significantly differentially expressed in $S_{F_i}$ samples compared to $\overline{S_{F_i}}$. Each bicluster may consist of exclusively over- or under-expressed features or can be of mixed-type and include features diverging from the background in both directions (Figure 2b).
UnPaSt identifies differentially expressed biclusters in a two-dimensional matrix in three subsequent steps: (i) feature binarization and feature selection, (ii) feature clustering (iii) sample clustering (Figures 2c-e).

In the first step, for each standardized feature $f_j$ (e.g., gene expression), all samples are split into two subsets with higher and lower values (Figure 2c). Unlike some other methods setting a fixed threshold[41,46], UnPaSt employs clustering to optimize the separation of the samples into two groups, $S_{f_j}$ and $\overline{S_{f_j}}$ representing bicluster and background sample sets (Figure 2f,g). Bicluster and background sets are defined such that $|S_{f_j}| < |\overline{S_{f_j}}|$, or assigned randomly when $|S_{f_j}| = |\overline{S_{f_j}}|$. Currently, users can choose between 2-means clustering, hierarchical clustering, and a mixture of two Gaussians, converting each feature into a binary vector where ones are assigned to samples from the bicluster set $S_{f_j}$. Then, UnPaSt selects features defining well-separated sample subgroups of at least $n_s$ samples ($n_s = 5$ by default), and thus indicating molecular heterogeneity, potentially being most relevant for patient stratification. The quality of the sample set $S_{f_j}$ separation from the background $\overline{S_{f_j}}$ based on $j$-th feature values is measured by the signal-to-noise ratio (SNR):

$$SNR(f_j, S_{f_j}) = \frac{|\mu_{f_j,S_{f_j}} - \mu_{f_j,\overline{S_{f_j}}}|}{\sigma_{f_j,S_{f_j}} + \sigma_{f_j,\overline{S_{f_j}}}},$$

where $\mu_{f_j,S_{f_j}}$ and $\mu_{f_j,\overline{S_{f_j}}}$ are mean values, and $\sigma_{f_j,S_{f_j}}$ and $\sigma_{f_j,\overline{S_{f_j}}}$ are standard deviations of feature $f_j$ in bicluster $S_{f_j}$ and background $\overline{S_{f_j}}$ sample sets, respectively.

SNR values computed for all features cannot be directly compared because the variance of estimated SNRs depends on the sizes of $S_{f_j}$ and $\overline{S_{f_j}}$ which vary across features. Therefore, to distinguish between well- and poorly binarized features, UnPaSt compares each observed $SNR(f_j, S_{f_j})$ with the distribution of SNR values obtained by splitting a standard normal distribution into two groups of the same size as $|S_{f_j}|$ and $|\overline{S_{f_j}}|$ $max(10000, \frac{10}{p})$ times (where $p$ is a user defined p-value threshold). Based on this null model, empirical binarization p-values are assigned to all features, and only features with p-values exceeding a user-defined threshold are passed to the second phase of the workflow (Figure 2g).

In the second step, binarized features defining well-separated sample sets are clustered into modules based on the similarity of their binary profiles, i.e., the resulting modules include features that distinguish similar sets of samples (Figure 2d). Currently, UnPaSt uses

WGCNA[47] with the "signed-hybrid" adjacency for feature clustering, which computes a similarity network for the binarized features, converts it into a topological overlap matrix, and selects a cutoff resulting in a network topology best fulfilling the scale-free property[48]. Alternatively to WGCNA, UnPast provides functionality for the users to specify their preferred similarity measure and clustering method, e.g., Louvain[49] or Leiden[50] algorithms maximizing modularity[51] when dissecting pairwise similarity networks into clusters. The feature clustering step is performed either for all features that passed binarization at once or separately for the two groups of features that define sample sets with under- or over-expression. Although the first approach allows finding mixed biclusters and “rescuing” otherwise not clustered individual features that are anti-correlated with all other features in a bicluster (like e.g., transcriptional repressors), UnPaSt demonstrated slightly higher performance when detecting over- and under-expressed biclusters independently.

In the third step, differentially expressed biclusters $B_i = (F_i, S_{F_i})$ are constructed from each module consisting of at least two features (Figure 2e). For that, samples are split into two clusters $S_{F_i}$ and $S_{\overline{F_i}}$ in a subspace of the feature set $F_i$ using the same binarization method used in the first step. As done for individual features, the SNR is computed for each bicluster by taking the average across features. As an optional post-processing step, UnPaSt applies *limma*[52] to each $B_i = (F_i, S_{F_i})$ to retain all features statistically significantly differentially abundant between $S_{F_i}$ and $\overline{S_{F_i}}$, given user-defined log2-fold-change and p-value thresholds.

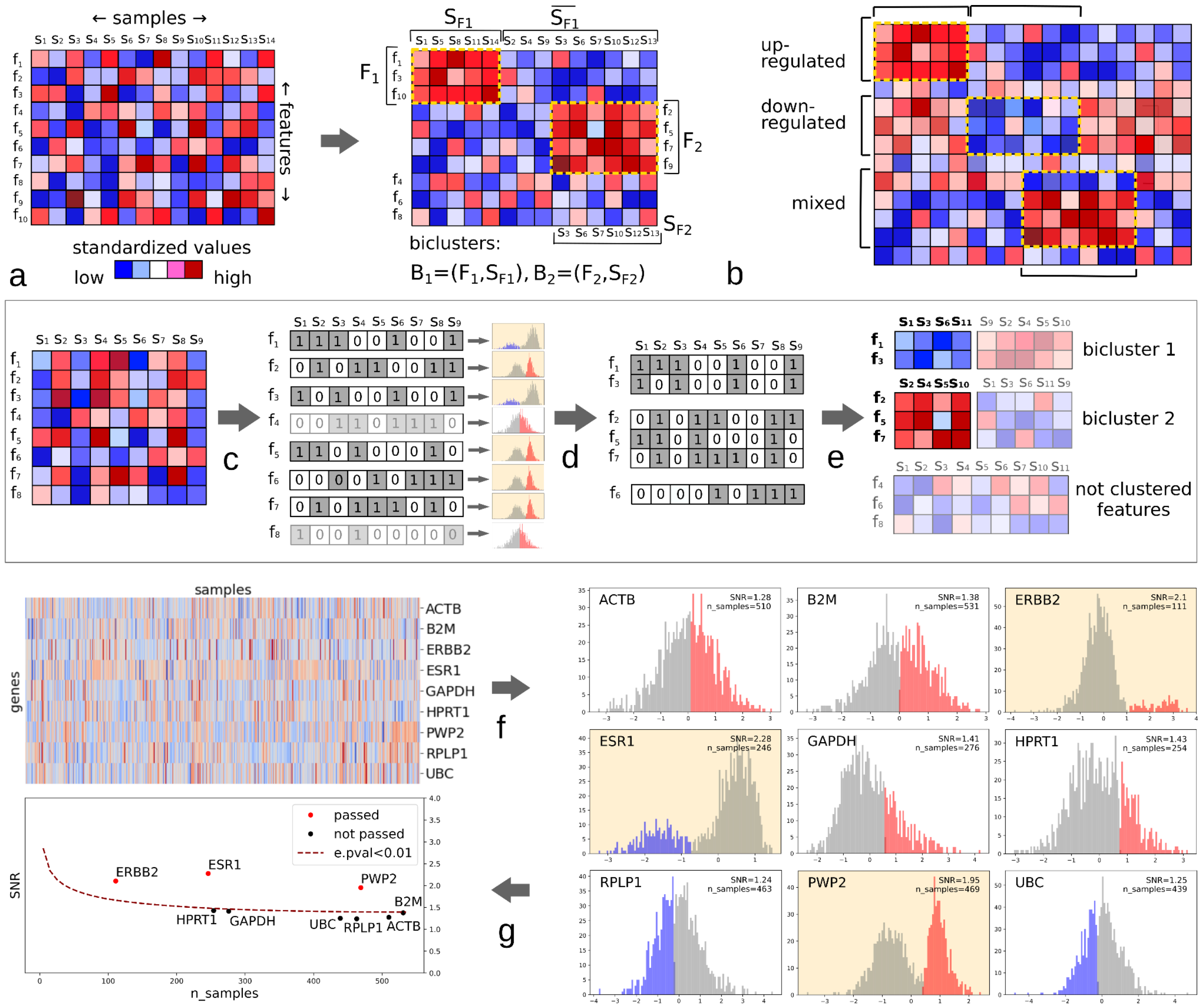

**Figure 2. a. Input and output of UnPaSt.** Given a two-dimensional matrix (on the left), UnPaSt identifies submatrices consisting of features and samples (on the right), such that these features are differentially abundant in corresponding samples. **b. Types of differentially expressed biclusters.** A bicluster can contain features that are either all over-expressed, all under-expressed, or represent a mix of both types of features. **c-e. The UnPaSt workflow** on the example of a matrix with two biclusters. **c. Feature binarization**. Each individual feature vector is turned into a binary vector where ones correspond to the smaller sample set. All features except $f_4$ and $f_8$, whose binarization gave two poorly separated sample groups with low SNR, passed to the next phase. **d. Clustering of binarized features. e. Sample clustering.** To construct biclusters, samples are split into two subsets based on the subset of features corresponding to each module. This results in two biclusters: one down-regulated (including $f_1, f_3$ and samples $s_1, s_3, s_6, s_{11}$) and one up-regulated bicluster (consisting of features $f_2, f_5, f_7$ and samples $s_2, s_4, s_5, s_{10}$). Three features $f_4, f_6, f_8$ remain unclustered. **f-g. The approach to feature binarization and feature selection** on the example of nine genes from the TCGA-BRCA dataset. **f.** Expressions of housekeeping genes ACTB, B2M, GAPDH, HPRT1, RPLP1, and UBC demonstrate bell-shaped distributions, which are binarized approximately by their medians. In contrast, known breast cancer biomarkers ESR1, ERBB2, and PWP2 show bimodal or heavy-tailed distributions and define two well-distinguishable sample sets each. The smaller sample subset is highlighted with red or blue, depending on whether it over- or under-expresses this gene compared to the background (larger sample set, colored gray). **g.** SNRs computed for each of the nine features are compared with the empirical distributions of SNRs obtained by binarizing standard normal distributions. Only ESR1, ERBB2, and PWP2 pass the empirical p-value threshold of 0.01 (red dashed line).

# Benchmark

## Simulated data

Since real data may contain many biological and technical patterns unrelated to disease, we first evaluated patient stratification methods using nine synthetic datasets, each consisting of 200 samples and 10,000 features and four simulated subtypes (Methods and Supplementary Figure S1). To investigate how data properties can influence the method performances, we simulated three scenarios with growing complexity (A, B, and C) and varied the number of features discriminating each subtype (5, 50 or 500) in each scenario. In scenario A, four non-overlapping clusters consisting of 5%, 10%, 25%, and 50% of all samples simulated four mutually exclusive disease subtypes. In scenarios B and C, sample clusters were allowed to overlap, i.e., each sample could be assigned to many clusters. In scenario C, to simulate additional variation unrelated to disease subtypes, four 500-feature co-expression modules not correlated with the four subtypes were added to each data matrix.

Because traditionally used clustering performance measures assume that the entire population is divided into disjoint sample subsets and are therefore not applicable to overlapping cluster structures, we computed the sum of weighted Adjusted Rand Indexes (ARI) for each simulated sample cluster and its statistically significant best match among all identified clusters or biclusters (wARIs; see Methods for details). Since the performance of the methods can largely depend on parameter selection[53], we have not only compared method wARIs with default parameters (Supplementary Figure S2), but also evaluated the methods with varying parameter combinations, and selected parameters resulting in the highest average performance across all nine datasets (Figure 3a, Supplementary Table S1b and Figure S3).

Across all scenarios, UnPaSt consistently demonstrated high wARIs, while the performance of most methods declined as the number of subtype-specific biomarkers decreased. When as few as five features discriminated between sample clusters, only several approaches could still identify sample groups that significantly overlapped with the true clusters. Also, clustering methods tended to perform worse when the clusters were overlapping and the data contained additional co-expression modules not related to sample clusters, as it is likely to happen in real-world omics data.

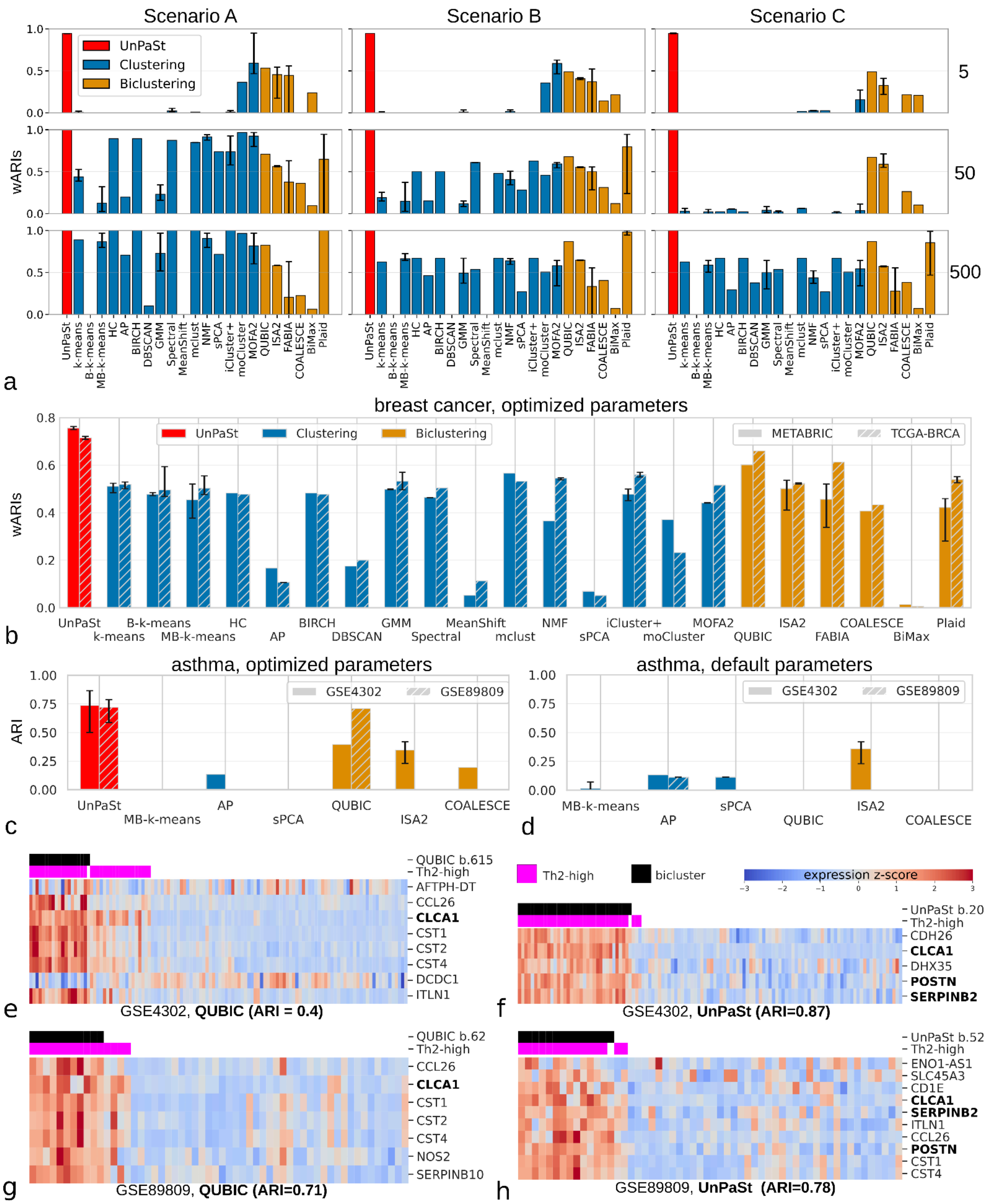

**Figure 3. a.** The performance of UnPaSt and other methods on simulated data, with parameters resulting in the best average wARIs over all nine datasets. For non-deterministic methods, bars represent the mean wARIs over five independent runs, and error bars indicate the range from minimum to maximum wARIs. **b.** The performances of patient stratification methods on METABRIC and TCGA-BRCA datasets, with the parameters having minimal average rank for both datasets. **c-d.** The results of *de novo* discovery of Th2-high asthma in the GSE4302[4] and GSE89809[54] datasets. Only methods demonstrating non-zero performances are displayed. Each method was applied with parameters optimized using breast cancer data (**c**) and with default parameters (**d**). **e-h.** Biclusters best matching the Th2-high asthma identified in the GSE4302 (**e**,**f**) and GSE89809 (**g**,**h**) cohorts by QUBIC (**e**,**g**) and UnPaSt (**f**,**h**). Since UnPaSt is a non-deterministic method, consensus biclusters from five independent runs are shown (see Methods for details). Each heatmap shows standardized expressions of bicluster genes (rows). Color bars above each heatmap specify sample membership in the bicluster (black) and whether the sample is over-expressing the Th2 signature (magenta) defined by three biomarkers CLCA1, POSTN, and SERPINB2 (highlighted with bold text font). In both datasets, UnPaSt biclusters match Th2-high asthma more precisely than QUBIC biclusters.

## Breast cancer

Breast cancer was one of the first diseases classified by gene expression, initially divided into four main molecular subtypes[55], now commonly assigned using supervised PAM50 classifier[56]. To assess the ability of UnPaSt and other evaluated methods to recover these subtypes, we applied them to the TCGA-BRCA[57] (n = 1089) and METABRIC[58] (n = 1904) datasets with varying parameter combinations (Supplementary Table S1b). Method performances were calculated as the wARIs for the PAM50-defined subtypes (Basal-like, Luminal, HER2-enriched, and Normal-like) and the identified sample clusters or biclusters.

While most methods showed substantial variability in performance across parameter settings, grid search led to only modest improvements compared to the results obtained with default parameters (Supplementary Figures S4). Notably, UnPaSt and ISA2 demonstrated consistently high wARIs across most of the tested parameters, which is advantageous in real-world scenarios where parameter selection is challenging.

To avoid overfitting for a specific dataset, we compared wARIs achieved with parameter combinations having the best average rank in TCGA-BRCA and METABRIC data across all combinations tested (Fig. 3b). Among all tested methods, UnPaSt demonstrated the highest wARIs (0.72 and 0.76 achieved by UnPaSt on average in five runs on TCGA-BRCA and METABRIC respectively, compared to 0.66 and 0.60 reached by the second-best method, QUBIC). Moreover, biclusters identified by UnPaSt in both datasets matched with sample sets defined as immunohistochemistry-confirmed overexpression of ER, HER2, and PR better than clusters or biclusters identified by any other method (Supplementary Figure S5). Compared to individual PAM50 subtypes, UnPaSt biclusters were the best matching for the Luminal and the second or third best matching for the Basal and HER2-enriched subtypes, slightly inferior to the findings of other biclustering methods. Normal-like, Luminal A and B, and Claudin-low subtypes were hardly detectable by all methods, possibly due to the lack of biomarkers discriminating them as isolated subgroups (Supplementary Figure S6). Nevertheless, sample groups best-matching Normal-like and Claudin-low subtypes were detected in both datasets by two other biclustering tools, ISA2 and COALESCE, respectively.

## Th2-high asthma

Asthma is a common inflammatory disorder of the respiratory system, and many studies report its symptomatic and molecular heterogeneity[59,60]. Woodruff et al. distinguished two molecular phenotypes of asthma based on the level of T-helper 2 (Th2)-driven inflammation in bronchial epithelium: “Th2-high” and “Th2-low” asthma[4]. Unlike breast cancer subtypes, only a few biomarkers differentiate Th2-high and -low asthma subtypes, making their unsupervised discovery a challenging task (Supplementary Figure S7).

To examine whether the optimization of parameters on breast cancer data generally improved method performance or only overfitted the methods specifically for breast cancer, we applied each method twice, with default parameters and the parameters optimized in the breast cancer benchmark.

Of all methods, only UnPaSt and QUBIC identified sample sets significantly and strongly overlapping with the Th2-high asthma defined as in Woodruff et al. (Methods) in both datasets GSE4302[4] and GSE89809[54] (Figure 3c-d). UnPaSt demonstrated the highest average ARI across five runs, followed by QUBIC with optimized parameters (GSE4302: mean 0.74 vs 0.40; GSE89809: mean 0.72 vs 0.71), which further increased to 0.87 and 0.78 for GSE4302 and GSE89809 by constructing consensus biclusters (Methods). Moreover, consensus UnPaSt biclusters best matching the Th2-high subset included all three biomarkers of Th2-high asthma (CLCA1, POSTN, SERPINB2), while the QUBIC biclusters included only CLCA1 (Figure 3e-h). When applied with default parameters, QUBIC and COALESCE produced no significant matches, and Affinity Propagation (AP), sparse PCA, and mini-batch k-means found significantly but weakly overlapping clusters with ARI <0.1 in GSE89809 and GSE4302, respectively (Figure 3d). No cluster or bicluster output by any other method overlapped the Th2-high group better than could be expected by chance (Fisher's exact test p-value cutoff of 0.05 after the Benjamini-Hochberg correction was applied).

## UnPaSt reveals molecular heterogeneity beyond established classifications

In addition to biclusters representing known disease subtypes, UnPaSt discovered numerous other patterns with comparable characteristics (size, SNR). These may represent either yet unrecognized disease subtypes or reflect variation unrelated to disease mechanisms. To prioritize biclusters more likely to reflect biologically meaningful expression patterns rather than false positives, batch effects, or contaminations, we sought those (i) recurring across independent datasets, (ii) associated with clinical variables and/or overrepresented with functionally similar genes.

### Asthma

Beyond the Th2 signature, UnPaSt revealed a pair of biclusters in asthma datasets that significantly overlapped in genes (adj.p-value < 0.05, Fisher’s exact test; Jaccard similarity 0.5; Supplementary Figure S8). Both biclusters were composed predominantly of sex chromosome genes (e.g. DDX3Y, EIF1AY, RPS4Y1, XIST) differentially expressed between males and females and perfectly separated samples by donor sex in GSE89809; no information about donors’ sex was available for GSE4302.

To further examine unmatched biclusters, we investigated their association with other available variables. Bicluster b.3 identified in the GSE4302 consisted of samples obtained from smokers (Fig. 4a; adj.p-value < 1.8e-14) and genes ALDH3A1, CYP1B1, and GPX2 previously reported to be up-regulated in response to cigarette smoke[61]. Although no direct gene-level matches for this bicluster were found in GSE89809, these genes were markedly overexpressed in the only sample from a current smoker in that dataset (Fig. 4b). This finding demonstrates how biologically meaningful patterns may appear irreproducible across datasets due to differences in sample composition rather than methodological limitations.

### Breast cancer

In the TCGA-BRCA and METABRIC datasets, UnPaSt identified 34 pairs of biclusters significantly overlapping in genes (adj. p-value < 0.05; Jaccard similarity > 0.1; ≥ 2 shared genes) and distinct from any of PAM50 subtypes (ARI < 0.25) in both datasets. Although none of these biclusters were significantly associated with overall survival in both datasets, they may reflect the aspects of breast cancer biology not encompassed by PAM50 classification.

To explore their potential biological roles, we tested genes from matched biclusters for overrepresentation of Gene Ontology (GO) and Disease Ontology (DO) terms, and for overlap with known pathways from KEGG and Reactome (Supplementary Table S2). Figures 4c-d show five expression signatures from bicluster pairs sharing 5-11 genes, which may reflect biologically relevant programs in breast cancer. Biclusters 100 and 163 (pink) were overrepresented with genes associated with lipid droplets (GO:0005811; adj.p-value=7.41e-8) and obesity (DOID:9970; adj.p-value=2.46e-07) and potentially reflecting the presence of adipocytes and adipogenesis, which have a known role in breast cancer[62]. Biclusters b.64 and b.35 (orange) contained exclusively genes from chr16q, a region frequently deleted in breast cancer[63]. Biclusters b.128 and b.60 (green) included interferon response genes (e.g. BATF2, IFIT1, ISG15; R-HSA-909733: adj.p-value=3.54e-25), and biclusters b.36 and b.22 (yellow) were enriched with genes associated with prostatic hypertrophy (DOID:11132; adj.p-value=2.03e-05), including many cancer/testis antigens (CSAG1, CSAG2, CTAG2 and MAGE family members). Both signatures potentially reflect specific features of the intratumoral immune landscape and may inform therapy selection and development[64,65].

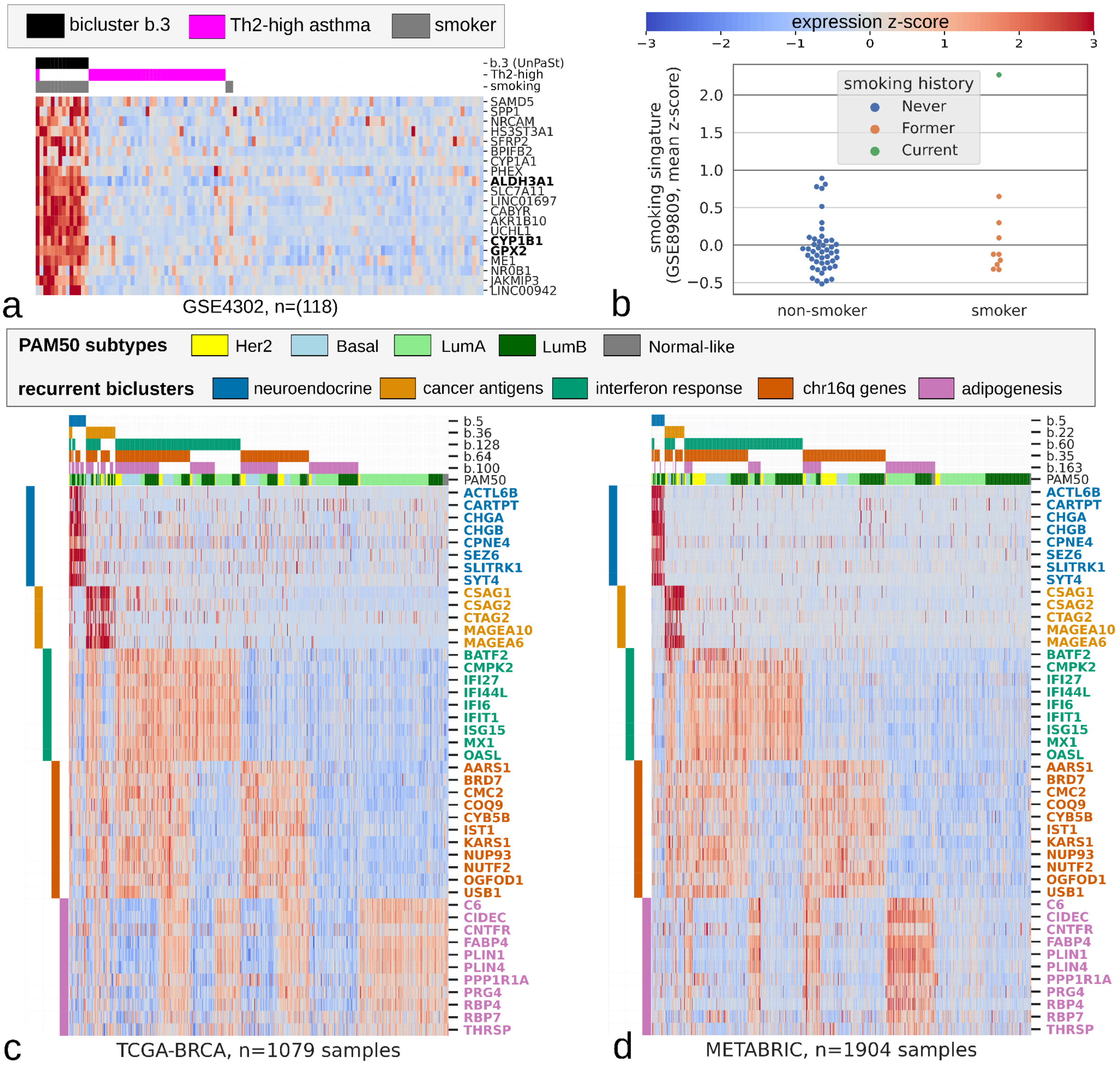

**Figure 4.** The examples of similar biclusters found by UnPaSt in independent datasets. **a**. A 20-gene bicluster found in the GSE4302 dataset with overrepresentation of samples obtained from smoking individuals (shown with the gray bar; only two out of 16 samples from smokers were not included into this bicluster). **b**. Mean z-scores of 20 genes associated with smoking in the GSE89809 were computed for samples from the GSE4302 cohort. The only sample obtained from a current smoker overexpresses this 20-gene signature higher than any other sample in the cohort. **c-d**. Five pairs of similar biclusters found in TCGA-BRCA (**c**) and METABRIC (**d**) datasets by UnPaSt and matched based on the similarity of their gene sets. Matched biclusters are highlighted

with bars and text font of the same color. Only genes shared between matched biclusters are shown; the full lists of genes in these biclusters are provided in Supplementary Figure S8c-d and Table S2.

Finally, biclusters b.5 in both datasets (blue) were enriched with presynaptic genes (GO:0098793; p.adj=4.21e-07), including neuroendocrine differentiation biomarkers chromogranin A (CHGA)[66] and SEZ6[67]. These may correspond to neuroendocrine breast carcinoma (NEBC), a rare subtype comprising approximately 2–5% of invasive breast carcinomas[66]. While specialized therapies against NEBC are not yet part of standard clinical practice, emerging evidence suggests that *SEZ6* may represent a novel promising therapeutic target in neuroendocrine cancers[67].

### Transcriptional heterogeneity across cancers

To explore transcriptional heterogeneity across other cancers, we applied UnPaSt to each of 33 cancer transcriptome datasets from TCGA and searched for (i) biclusters from representing recurrent expression signatures across different cancer types, and (ii) biclusters potentially representing novel overall survival (OS) biomarkers. The recurrent signatures found in up to five cancer types were overrepresented with functionally coherent genes (e.g., sex-specific genes, genes overexpressed in lymphoid cells or liver tissue, and spliceosomal genes). However, some of these recurrent expression signatures also correlated with experimental batches, suggesting potential cross-sample contamination (Supplementary Text and Figures S9-S14).

Among the biclusters significantly associated with OS, 184 were dissimilar from previously defined RNA expression-based clusters. For in-depth investigation and validation of potential biomarker candidates identified by UnPaSt, we have chosen the dataset with the largest number of OS-associated biclusters, TCGA-KIRC, and confirmed the presence and survival association of two similar expression signatures in an independent microarray-based dataset E-MTAB-1980[68] (Supplementary Text and Figures S15-S16). Both signatures were predominantly found in samples classified as ccB[69] subtypes and obtained from patients with significantly shorter OS compared to the rest of the patients in both cohorts (Fig. 5a and 5b). KIRC_193 and its best match in E-MTAB-1980 shared IGF2BP3, TNNT1, and PTPRH genes, all previously linked with cell invasion and migration and reported to be associated with poor prognosis in various cancers[70–73]. KIRC_217 and its best match contained a tumor suppressor gene, EMX2, and its enhancer antisense RNA, EMX2OS, whose downregulation was previously associated with reduced survival in ccRCC[74,75]. Both signatures were also detected in the ICGC RECA-EU dataset and showed consistent associations with the ccA/ccB classification. However, none of the survival associations observed in TCGA and E-MTAB-1980 were replicated in RECA-EU, possibly due to cohort-specific clinical differences, limiting its utility as a validation dataset in survival analysis (Supplementary Figures S17).

## Applications to other data types and designs

While originally developed for patient stratification using bulk transcriptomics data, UnPaSt has proven itself useful for the analysis of other diverse datasets, including longitudinal profiles of deletion-containing viral genomes (DelVGs)[76], temporal changes in the cerebrospinal fluid (CSF) proteome in spinal cord injury (SCI)[77], breast cancer multi-omics[2,78], single-cell transcriptome of PBMC[79], and renal spatial transcriptomics (ST) data[80].

## Longitudinal data

In both longitudinal datasets, UnPaSt identified biclusters corresponding to temporal dynamics of biologically meaningful patterns. Specifically, it extracted several clusters of DelVGs with concordant longitudinal profiles, different from clusters identified by spectral co-clustering, and revealed a protein expression signature that may be an early biomarker of SCI severity (Supplementary Text and Figures S18-S19).

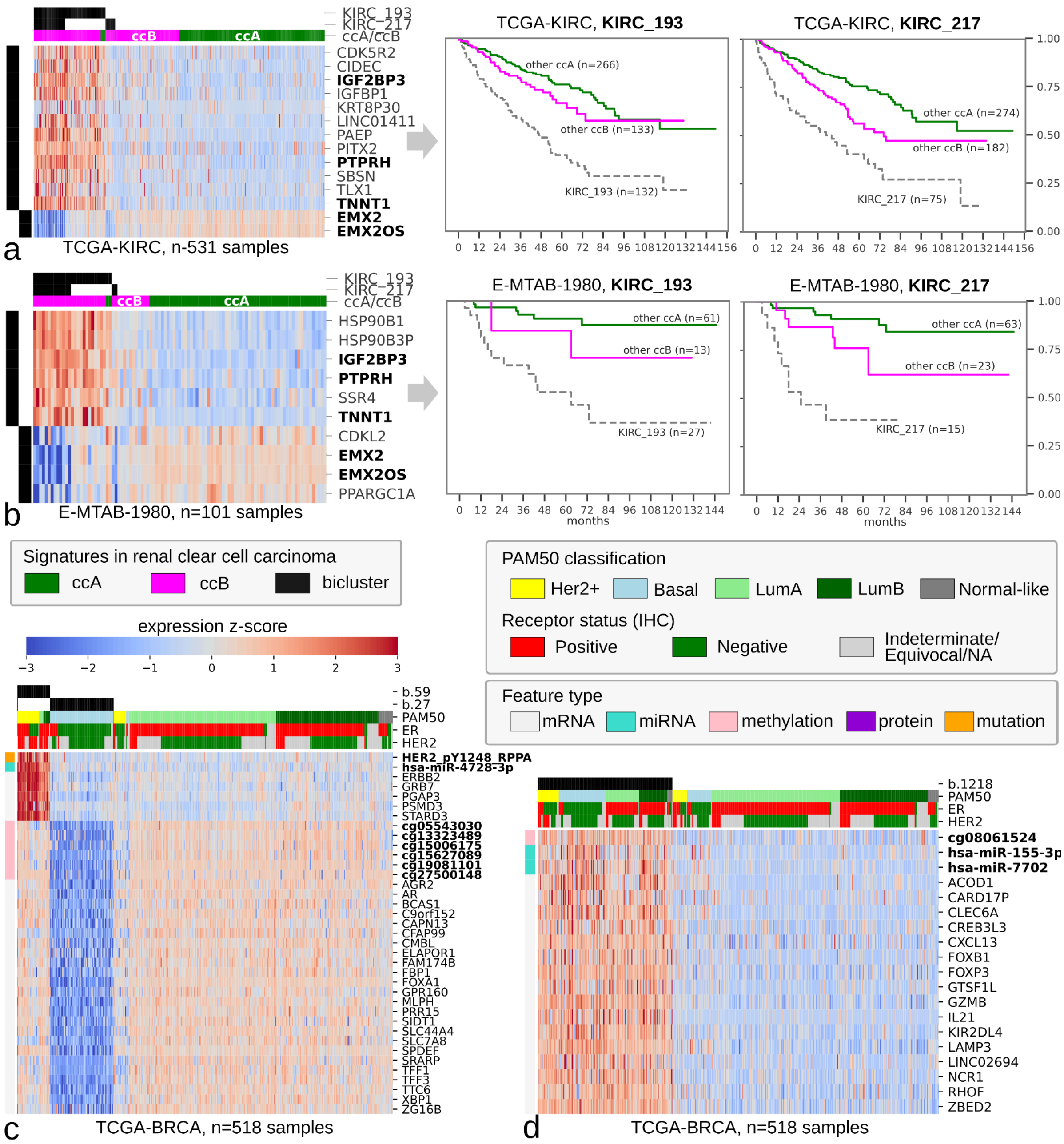

**Figure 5. a-b.** OS-associated biclusters KIRC_193 (up-regulated) and KIRC_217 (down-regulated) found in the kidney cancer dataset TCGA-KIRC[81] (**a**) and their best matching biclusters found in E-MTAB-1980[68] (**b**). Sample classification into ccA and ccB molecular subtypes[69] is shown with green and magenta labels on the tops of the heatmaps. Shared genes are highlighted with bold text font. **c.** Multi-omics biclusters best match Her2+ (top) and Basal-like subtypes (bottom) defined by the PAM50 classifier. Features corresponding to omics types other than mRNA expression are highlighted with bold text font. Bicluster 59 includes HER2 protein expression (HER2_pY1248_RPPA) and miRNA miR-4728-3p in addition to the expression of ERBB2. This bicluster seems to match the subset of samples with IHC-confirmed Her2 overexpression even more precisely than the Her2+ subtype defined by the PAM50 classifier (yellow). **d.** Bicluster 1218 does not match the PAM50 subtypes and includes methylation and miRNA features, in addition to expressions of genes specific to NK and T cells (e.g., KIR2DL4, IL21, NCR1, GZMB, FOXP3), potentially indicative of invasion of these cells in tumors.

## Multi-omics data

In a multi-omics dataset obtained from XENA[78] and encompassing five modalities (mRNA, miRNA, and protein expression, methylation, and somatic mutation data) and almost 30,000 features across 518 breast tumor samples, UnPaSt identified 1398 biclusters, including both well-matching known subtypes (Fig. 5c), and those independent from PAM50 classification (Fig. 5d). Notably, although only 31 (2.2%) biclusters consisted of features from multiple omics types, the biclusters best matching Basal-like, Her2+, and Luminal subtypes included features from multiple omics types. These findings support the notion that cancer subtypes are discernible across multiple omics layers, although their signatures may be more pronounced within individual modalities[24,25,38,82]. Concurrently, biclustering analysis indicates that a large proportion of molecular heterogeneity within this dataset is specific to individual omics modalities beyond mRNA and warrants further exploration (Supplementary Text and Figures S20).

## Single cell and spatial data analysis

In the 3k PBMCs dataset[79], comprising 2638 cells from a healthy donor, UnPaSt identified biclusters distinguishing previously defined cell types (Fig. S21), including very rare ones, like megakaryocytes (15 cells, 0.57%) and dendritic cells (37 cells, 1.4%). In addition to revealing similarities and dissimilarities between cell populations of myeloid and lymphoid (Figs. 6a and S22) lineages, UnPaSt detected biclusters corresponding to biologically relevant cell subpopulations, such as resting CD4+ T cells (Fig. 6b, bicluster b.1028) overexpressing CCR7, LEF1, and TCF7, or FCRG3A+ NK cells and monocytes (Fig. 6b, bicluster b.1015; and S22), and biclusters reflecting activation of broad transcriptional programs across multiple cell types, e.g., the AP1 pathway (Fig. 6b, bicluster b.1020).

In a spatial transcriptomics dataset[80] comprising 10,763 spots from 19 renal biopsy slides obtained from patients with antineutrophil cytoplasmic antibody-associated glomerulonephritis (ANCA-GN), UnPaSt identified 187 biclusters. Of these, twelve included marker genes used by Engesser et al. to define renal compartments by interpreting non-overlapping clusters of spots[80] (Supplementary Text, Table S7a, and Figure S23). Notably, some of these biclusters demonstrated strong spatial overlaps, for example biclusters b.126 and b.139, corresponding to the thick ascending limb (TAL) and thin limb of the loop of Henle (LOH), respectively (Fig. 6c). Given the spatial proximity of these compartments within the nephron, we hypothesize that certain spots may have overlapped multiple compartments.

Furthermore, among the 187 biclusters identified in the ANCA-GN group, 15 were not observed in the control samples (3 ST slides, 9,065 spots total) and potentially represented disease-specific expression patterns. Many of these biclusters consisted of genes specific to various immune cell subpopulations (T cells, B cells, macrophages) or interferon response, suggesting that immune response appears to manifest in many spots not previously designated as inflamed (Fig. 6d-e, Supplementary Text, Table S7b and Figure S24). Notably, the prevalence and spatial distribution of these disease-associated biclusters varied widely across slides and slices (Fig. 6e, Supplementary Figures S25, S26). Thus, UnPaSt enabled a more nuanced decomposition of ST data compared to conventional clustering. It detected not only the main renal compartments, but also identified patterns potentially indicative of disease-specific processes and revealed their spatial heterogeneity across individuals.

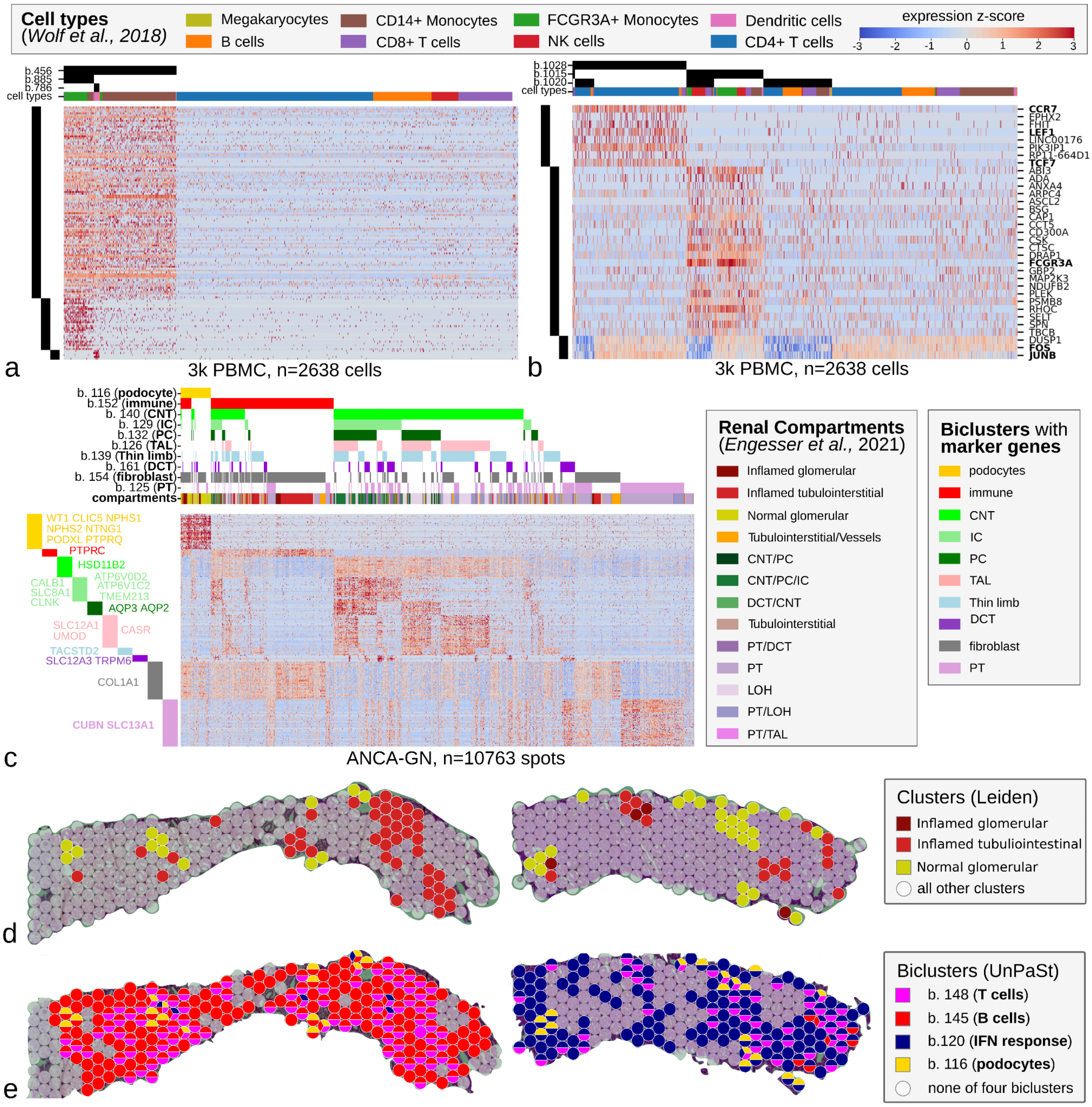

**Figure 6. a-b.** Biclusters identified by UnPaSt in the 3k PBMC single cell dataset. Three biclusters best matching all cells of myeloid origin (b.456), and distinguish FCRG3A/CD16+ monocytes (b.885) and dendritic cells (b.786) from classical monocytes. Names of biomarkers included into each bicluster are displayed on the left side. **b**. Three biclusters corresponding to activation of specific transcriptional programs within and across cell types. Bicluster 1028. Bicluster 1015 overexpresses FCRG3A gene and consists mainly of cells previously classified as NK or FCRG3A+ non-classical monocytes. Bicluster 1020 consists of AP-1 pathway genes and potentially reflects its varying activity across all cell types.

**c**. Ten biclusters matched to renal tissue compartments previously defined using Leiden clustering and annotated based on expression of marker genes (immune - immune cells, CNT - connecting tubules, IC - intercalated cells, PC - principal cells, LOH - loop of Henle, TAL - thick ascending limb of LOH, Thin limb - thin ascending limb of LOH, DCT - distal convoluted tubules, PT - proximal tubules). Biclusters are highlighted with color bars on the edges of the heatmap and are matched by colors with marker gene sets they overlap. Marker genes included into the biclusters are shown on the left side of the heatmap. Up to 50 gene expressions per bicluster are displayed, two biclusters with PT and PC biomarkers are not shown. **d-e.** The spots of the ST slide V6_A with two slices obtained from individual with ANCA-GN colored according to their assignment to (**d**) three compartments defined by Leiden clustering[79,83] and to four biclusters (**e**) with markers of T cells (b.148, magenta), B cells (b.154, red), interferon response (b.120, dark blue), and podocytes (b.116, yellow). Spots belonging to multiple biclusters are displayed as pie charts with 2-4 equal-sized sections. Spots not assigned to any of these four biclusters are displayed as white circles. Biclustering analysis reveals large differences in immune response profiles in two sections from different regions of the same kidney biopsy specimen.

# Discussion

In this study, we benchmarked 23 methods for unsupervised patient stratification using multiple real and simulated datasets. Our results demonstrate that most stratification methods achieve high performance when sample clusters are well-differentiated and non-overlapping. In such scenarios, computationally complex data integration methods, whose popularity has been growing in recent years, appear similar or slightly inferior in terms of performance compared to conventional clustering algorithms, while requiring substantially more runtime and memory (Supplementary Text). In more challenging settings, where true sample groupings are defined by a small number of features and obscured by other patterns, conventional clustering methods substantially underperform.

To eliminate this deficiency, we developed UnPaSt, a biclustering-based method for unsupervised patient stratification. Unlike its competitors, UnPaSt was efficient in all scenarios and detected patterns of various kinds, from frequent and strongly differentially expressed signatures defining Luminal and Basal breast cancer subtypes to smaller signatures consisting of only several biomarkers (e.g., Th2-high asthma, sex-specific gene expression, and smoking signature), or signatures corresponding to rare disease subtypes (e.g. breast cancer with neuroendocrine differentiation). This sensitivity to patterns manifesting in a small subset of all features makes UnPaSt a promising tool for discovery of disease subtypes characterized by a few specific biomarkers. Ultimately, exploring disease heterogeneity at higher resolution can facilitate the discovery of biomarkers and inform the development of novel precision therapies.

The performance of UnPaSt in real and synthetic benchmarks is attributed to two factors: (i) it reduces the patient stratification problem to a biclustering problem, which enables the discovery of overlapping sample sets, and (ii) its feature selection procedure aims specifically at feature subspaces, in which the whole sample cohort can be divided into two distinct subsets with high and low abundance of corresponding features. In other words, the problem solved by UnPaSt can be viewed as an unsupervised version of differential expression analysis problem, where target classes are not known *a priori* and are to be found. Although UnPaSt does not guarantee that the discovered bicluster features exhibit statistically significant differential abundance, this hypothesis can be tested after bicluster detection, using any state-of-the-art differential expression analysis methods, such as *limma*[52].

Besides evaluation of UnPaSt using transcriptomic datasets with known molecular subtypes, we demonstrated its utility for analyzing many other types of biological data, including multi-omics, single-cell, and spatial data. We have shown that UnPaSt goes beyond detecting the largest non-overlapping patterns visible for traditional clustering methods, yielding a more detailed overview of data heterogeneity. This not only makes UnPaSt a promising alternative to popular clustering and data integration tools but also highlights perspectives for its potential application as a feature selection and dimensionality reduction technique. In the first step, UnPaSt extracts features delineating two distinct sample subpopulations and excludes the rest. These features represent promising biomarker candidates and are potentially more informative for training robust classifiers or other machine learning models. Further, UnPaSt assembles biclusters by grouping features that define the similar partitions of all samples into two subgroups with an opposite differential expression pattern, which can be encoded by a binary variable. In our experiments, UnPaSt typically detected hundreds of biclusters in datasets with tens of thousands of features, suggesting that it can potentially reduce data dimensionality by up to two orders of magnitude, but its added value in comparison to popular dimensionality reduction techniques remains to be determined.

Besides applying UnPaSt to other unsupervised learning problems, expanding the benchmark to additional datasets with validated subtypes and signatures and to a broader method portfolio, including exploring how novel AI-based techniques could enhance unsupervised stratification, represent promising directions for future research. Beyond that, we also see substantial potential for further method improvement, for example, by replacing the current binarization step with discretization. This may enhance performance on highly heterogeneous data, where features can define three or more well-separable subgroups. The second step, which currently relies on WGCNA for feature clustering, could also be improved by exploring alternative approaches, e.g., community detection algorithms such as Leiden[50]. To reduce the chance of capturing false positives, UnPaSt applies an empirical p-value cutoff in the first step and considers only biclusters with at least two features.

While these measures can help to reduce false positives, direct estimation of error rates using real biological data remains a challenging task, since we lack complete knowledge of which patterns are actually present in the real datasets. Additionally, to eliminate likely dataset-specific technical patterns, in this work, we focused on biclusters that replicated in independent datasets and/or were significantly associated with clinical (and not technical) variables. Nevertheless, independent validation of the discovered patterns in external datasets and experimental studies will remain essential to link the identified biclusters to true biological mechanisms.

# Supplementary data

# Methods

## Simulated data

To analyze the behavior of the methods under different circumstances, we defined three scenarios with increasing complexity and varied the number of subtype-specific features from 500 to 50 and 5 in each scenario. This resulted in nine 10000-feature matrices (3 scenarios x 3 feature set sizes) simulating omics data derived from a cohort of 200 patients classified into four subtypes. Samples of each dataset were randomly assigned to four subtypes, including 10, 20, 50, or 100 samples. In scenario A, simulated subtypes were mutually exclusive, i.e., each sample belonged to no more than one subtype. In scenarios B and C, samples were assigned to each subtype independently, and subtypes overlapped in samples, i.e., each sample was assigned to 0-4 subtypes. First, each matrix was filled with values drawn from a standard normal distribution simulating the background. Second, four non-overlapping sets of $n = 5, 50, 500$ features were randomly chosen to represent subtype-specific biomarkers. Third, for each subtype, values corresponding to subtype-specific genes were replaced with the values drawn from $\mathcal{N}(4, 1)$. Finally, in scenario C, four co-expression modules not associated with target subtypes and containing 500 features were added to the background. To add each co-expression module, we sampled 500 background features and modified all feature vectors $f_i$ except the first one as $f_i' = f_0 r + f_i\sqrt{1-r^2}$, where $i = 1 - 499$, and $r = 0.5$ to establish Pearson's correlations between features from one module to be around 0.25.

## Real data preprocessing

RSEM-normalized and log2-transformed gene-level read counts, exon-level expressions, and sample information for the TCGA-BRCA cohort[3] were obtained from XENA[78] TCGA Pan-Cancer (PANCAN) data hub (https://xenabrowser.net/datapages/?cohort=TCGA%20Pan-Cancer%20(PANCAN)). Out of 1108 samples in the TCGA-BRCA cohort, 1079 samples were obtained from primary tumors of females and 29999 genes with at least 3 normalized read counts in at least 5 samples were kept. Normalized and log2-transformed expressions of 24368 genes in 1904 samples and clinical data from the METABRIC cohort[58] was downloaded from cBioPortal [84,85] (https://www.cbioportal.org/study/summary?id=brca_metabric). To simplify the comparison and validation of findings in METABRIC and TCGA-BRCA, 17162 genes presented in both datasets were kept.

Raw .CEL files and sample annotations for GSE4302[86] and GSE89809[54] datasets were obtained from the NCBI GEO[87] database (https://www.ncbi.nlm.nih.gov/geo/). Only bronchial epithelium samples were included in the analysis. Raw files were downloaded, read, background-corrected, RMA-normalized, and log2-transformed using R packages *GEOquery* 2.54.1[88], *affy* 1.64.0 and *affyio* v1.56.0[89]. Probe IDs were mapped to gene names using *biomaRt*[90] v2.42.1 and probes with maximal variance were selected per gene using *collapseRows()* function from the *WGCNA*[47] v1.70-3 R package.

The acquisition and preprocessing of multi-omics, proteomic, delVGs, single-cell, and spatial transcriptomics data are described in detail in the Supplementary Text.

## Identification of known molecular subtypes in breast cancer and asthma

Unlike in the TCGA-BRCA dataset, in METABRIC the claudin-low subtype was treated as the sixth mutually exclusive subgroup not overlapping with any PAM50 subtype, to unify sample classifications in TCGA-BRCA and METABRIC datasets. Therefore, we reassigned subtype labels using supervised classifiers from the *genefu* (*https://bioconductor.org/packages/release/bioc/html/genefu.html*) R package v2.18.1. Expression profiles from TCGA-BRCA and METABRIC datasets were classified into Luminal A, Luminal B, HER2-enriched, and Normal-like subtypes using the *molecular.subtyping()* function with sbt.model = "pam50". In agreement with *Fougner et al.*[91], we considered claudin-low as an additional phenotype independent from PAM50 classification and subdivided all samples into claudin-low merged with PAM50 classification and non-claudin-low subsets regardless of their PAM50 labels using *claudinLow()* function. Luminal, Basal, and HER2-enriched subtypes predicted by supervised classifiers closely resembled published classifications, while the assignment of samples into Luminal A and B subsets in both datasets, and into Normal-like in METABRIC differed (Supplementary Figure S7a-b).

Th2-high asthma subgroups in GSE4302 and GSE89809 cohorts were identified using the same approach as described by Woodruff et al[4]. In each cohort, samples were separated into two groups based on the expressions of three biomarkers, POSTN, SERPINB2, and CLCA1 using hierarchical clustering (Python *scikit-learn* v.1.2.2) with complete linkage and Euclidean distance. This resulted in 37 out of 118 (31%) and 15 out of 56 samples (27%) classified as Th2-high in GSE4302 and GSE89809 cohorts, respectively.

### Methods and parameter settings

We used *scikit-learn* v.1.2.2 for clustering methods Affinity Propagation, Hierarchical Clustering, Bisecting k-means, BIRCH, DBSCAN, GMM, k-means, MeanShift, Mini-batch k-means, Spectral, *MoSBi*[92] v1.10.0 R package for biclustering algorithms BiMax, FABIA, ISA2, Plaid, and QUBIC, and JBiclustGE-CLI[93] v1.0.0 for COALESCE, and UnPaSt implementation available at https://github.com/ozolotareva/unpast_paper/. The full list of evaluated methods, links to their publicly available implementations, software versions, and tested parameters is provided in Supplementary Table S1. Since our previously published results demonstrated that the performance of many biclustering tools can be largely improved by proper parameter selection[53], in this work, we used grid search to tune methods' parameters on simulated and real breast cancer data. In real-world conditions, however, parameter tuning by optimizing an external performance metric cannot be implemented due to the absence of complete ground truth. Therefore, this work does not provide the recipes for parameter tuning but rather evaluates the influence of parameters on the methods' performances.

## Performance evaluation

Although molecular classifications are established for some diseases, evaluating patient stratification results remains a challenging task. Among the approaches used in previous studies[21–26], three strategies can be distinguished: direct estimation of method performances through the comparison of the predicted subtypes with the known ones by (i) using simulated data, (ii) using real data, or (iii) assessing method performances using proxy measures, such as association between predicted subtypes and clinical features. However, all three approaches might give biased performance estimates. Superior method performance in a synthetic data benchmark does not necessarily guarantee that this method will work equally well in more complex real-world scenarios[53]. Correctly identified molecular subtypes do not always correlate with available clinical features such as survival or drug response. Real data might contain many biologically meaningful patterns not related to known disease subtypes (e.g., unknown disease subtypes, sex-specific gene expression) or experimental artifacts (e.g., batch effects, contaminations), hindering the discrimination between actual true and false positives. Moreover, traditional clustering performance measures, such as ARI or mutual information, assuming that the entire population of patients is divided into disjoint sample subsets, are not applicable for evaluation of biclustering results that contain many overlapping sample clusters. Additionally, the average number of clusters and the distribution of cluster sizes widely vary across evaluated methods, spanning the range from several clusters to thousands. Therefore, to minimize the contribution of predicted clusters that match known clusters just by chance, it is necessary to consider the statistical significance of the observed matches between predicted and known clusters. To enable performance estimation for overlapping sample clusters and to reduce the influence of unaccounted heterogeneity on performance estimates, we calculated the sum of weighted ARIs (wARIs) computed for each known subtype and its statistically significant best match among all identified sample clusters or biclusters, identified as follows. Let $S$ be a set of all samples in the cohort, and clustering $C_{true} = \{C_{t_1}, ..., C_{t_m}\}$ represents an established molecular classification with $m$ disease subtypes that can overlap. To identify the best matches of each known subtype $C_{t_i} \in C_{true}$ among $n$ predicted sample clusters $C_{pred} = \{C_{p_1}, \ldots, C_{p_n}\}$, the following three steps are performed.

1. For each pair of known and predicted clusters $C_{t_i}$ and $C_{p_j}$, the statistical significance of the overlap is assessed with Fisher's exact tests and the lowest two-tailed p-value adjusted for multiple testing using the Benjamini-Hochberg method is stored.

2. For each ground truth cluster $C_{t_i}$, the predicted clusters which most specifically match $C_{t_i}$ and not any other true cluster $C_{t_{k \neq i}} \in C_{true}$ (i.e. having the lowest p-value) are considered to be the best match candidates of $C_{t_i}$.

3. For each of the best matching candidates of $C_{t_i}$ where the adjusted p-values remain below the user-defined threshold (e.g., 0.05, as set in this work), the cluster $C_{p_j}$ with the highest ARI with $C_{t_i}$ is chosen as its best match.

When best matches are identified for all $C_{t_i} \in C_{true}$, overall performance is computed as the sum of their ARI weighted proportionally to the size of $C_{t_i}$:

$wARIs = \sum_{i}^{m} ARI(C_{t_i}, C_{p_j}) w_{t_i}$, where $w_{t_i} = \frac{|C_{t_i}|}{\sum_{i}^{m} |C_{t_i}|}$, and $C_{p_j}$ having the highest ARI across all candidates from $C_{pred}$ significantly overlapping $C_{t_i}$.

## Consensus biclusters

Since UnPaSt is a non-deterministic method, its results obtained in $n$ independent runs with the same input and parameters may vary. To obtain a more stable result and remove poorly reproducible patterns detected only in individual runs, we combined biclusters obtained in multiple independent runs with the same parameters into consensus biclusters as described below.

1. For every pair of bicluster sets, the pairs of significantly best-matching biclusters were identified based on their sample set similarity following the same procedure as was used for matching predicted sample clusters with ground truth and described in the *Performance evaluation* section above.

2. Pairwise similarity matrix for all biclusters detected in $n$ runs was filled with sample-based Jaccard similarities for the best matching bicluster pairs or zeroes for all other pairs. Louvain[49] clustering was performed to identify sets of best matching biclusters identified in independent runs. The elbow method was applied to select an optimal similarity cutoff between 0.33 and 0.9.

3. For each group of matched biclusters, the consensus gene set is defined based on the frequencies of each gene appearance. Only biclusters found in at least $\rho$ of all runs were retained and genes that were included in at least two biclusters were kept. We set $\rho = \frac{1}{3}$ to keep biclusters found in at least two out of five independent runs.

4. For each consensus gene set consisting of at least two genes, samples are divided into bicluster and background subsets using the same approach as for feature binarization in the first step.

## Statistical analysis

The statistical significance of overlaps between pairs of gene or sample sets was assessed using Fisher's exact test. For the evaluation of bicluster overlaps, the Chi-squared test was used. The Python package scikit-learn[94] (version 1.3.2) was utilized for the computation of test statistics and p-values.

To evaluate the association of a sample cluster with survival, the Cox Proportional Hazards model was employed, adjusted for the donor's age at the time of diagnosis, sex, and tumor stage (where available). Sample membership in a cluster or bicluster, stage, and sex were modeled as binary variables. Time-to-event analysis and plotting of Kaplan-Meier curves were performed using the Python package lifelines (version 0.25.10)[95].

Gene set overrepresentation analysis was performed using clusterProfiler[96,97] R package (version 3.14.3) and gene sets comprising 5-500 genes from GO[98,99], DO[100], and KEGG[100,101] and Reactome[102] databases. In each test, the set of all expressed genes annotated in the database was used as the background. Overlaps passing an adjusted p-value cutoff of 0.05 were considered significant and only overlaps including at least two genes were taken into account.

Differential expression analysis was performed using *limma* 3.58.1 (or *limma-voom* for count data)[52], upper quartile normalization[103] from *edgeR* 4.0.16[96]. Genes with |logFC| >1 passing an adjusted p-value cutoff of 0.05 were considered to be differentially expressed.

In all cases of multiple testing, the Benjamini-Hochberg procedure[104] was performed, except for the evaluation of bicluster redundancy analysis, where Bonferroni correction was applied instead.

## Code availability

The code used to reproduce the analyses and figures in this study is publicly available at https://github.com/ozolotareva/unpast_paper. A PyPI package containing development versions is available at https://pypi.org/project/unpast/. For user convenience, UnPaSt is also available through a web interface for datasets up to 200 MB at https://apps.cosy.bio/unpast/; the web frontend and backend code are available at https://github.com/UnPaSt/unpast-frontend and https://github.com/UnPaSt/unpast-backend, respectively.

# Acknowledgements

We thank Dr. Tim Daniel Rose for insightful discussions on consensus biclustering approaches.

This work was supported by the German Federal Ministry of Research, Technology and Space (BMFTR) within the framework of "CLINSPECT-M" (grant 161L0214A) (O.Z. and J.B.). This work was supported by the German Federal Ministry of Research, Technology and Space (BMFTR) within the framework of "CLINSPECT-M-2" (grant 03LW0243K) (Y.B. and J.B.). This work was funded by the Deutsche Forschungsgemeinschaft (DFG, German

Research Foundation) – 517063424 (J.-O. S. and J.B.). This work was developed as part of the PoSyMed project and is funded by the German Federal Ministry of Research, Technology and Space (BMFTR) under grant number 031L0310A (Z.C., O.T., and J.B.). Zoia Chervontseva was supported by (a postdoctoral fellowship from) the Peter und Traudl Engelhorn Stiftung. This work was developed as part of the DrugSiderAI project and is funded by the German Federal Ministry of Research, Technology and Space (BMFTR) under grant number 031L0306B (F.F. and J.B.). This work was funded by the European Union. Views and opinions expressed are however those of the author(s) only and do not necessarily reflect those of the European Union or the European Research Executive Agency. Neither the European Union nor the granting authority can be held responsible for them (F.M.D.-C. and J.B.). This work was also partly supported by the Swiss State Secretariat for Education, Research, and Innovation (SERI) under contract No. 22.00115 (F.M.D.-C. and J.B.). F. M. S. P. was supported by BAYER Foundation 2022 Carl Duisberg Fellowships for Medical Sciences. This work was supported by the German Federal Ministry of Research, Technology and Space (BMFTR) and the Universität Hamburg with funds of the Excellence Strategy of the Federal Government and the Länder (grants to J.J.G.L and T.L.). This work was supported by the German Federal Ministry of Research, Technology and Space (BMFTR) within the framework of the *e:Med* research and funding concept (*grants 01ZX1910D and 01ZX2210D* to T.L. and J.B.). K. V. S. was supported by FNRS convention PDR T.0294.24 “Expanded PRS embracing pathways and interactions for increased clinical utility”.

# Contributions

O. Z. conceived and led the project, co-supervised by J. B., M. E. and J. B. supervised early algorithm development. O. Z., A. S., M. H., A. A., and Y. B. developed and implemented the UnPaSt algorithm (excluding the webserver). Specifically, O. Z. implemented the first version of the core algorithm; M. H., A. S. and A. A. optimized the feature binarization step; M. H. and A. S. refactored the codebase to improve usability and installation; Y. B. created the Docker environment and PyPI package and added the function for differential expression analysis. M. H. and A. M. created the web server. M. H., A. M., Y. B., F. M. S. P., D. H., K. K., Z. C., N. P., O. T., A. A. tested UnPaSt across datasets. M. H., A. M., F. M. D.-C., J.-O. S., F. M. S. P., D. H., K. K., O. Z. performed benchmarking of patient stratification methods. M. H., A. M., Y. B., A. A., F. M. S. P., D. H., J. J. G. L., A. H., Z. C., F. F., K. S., M. K., V. G. P., O. Z. performed data preprocessing, analyses and visualizations. A. A. developed functions for better bicluster visualization. O. I. I., F. M. S. P., D. H., C. S., J. J. G. L., E. Z., M. K., V. G. P., T. L., and O. Z. provided biological interpretation of the results. O. Z., M. H., A. M., O. I. I., Y. B., D. H., J. J. G. L. prepared the initial manuscript draft; O. Z., M. E., J. B., K. V. S., V. G. P., T. L., C. S., D. B. B., E. Z., O. T. supervised the study. All authors critically reviewed and approved the final manuscript.

# References

1. Cho, J. H. & Feldman, M. Heterogeneity of autoimmune diseases: pathophysiologic insights from genetics and implications for new therapies. *Nat Med* **21**, 730–738 (2015).

2. Hoadley, K. A. *et al.* Cell-of-Origin Patterns Dominate the Molecular Classification of 10,000 Tumors from 33 Types of Cancer. *Cell* **173**, 291–304.e6 (2018).
3. Liu, J. *et al.* An Integrated TCGA Pan-Cancer Clinical Data Resource to Drive High-Quality Survival Outcome Analytics. *Cell* **173**, 400–416.e11 (2018).
4. Woodruff, P. G. *et al.* T-helper type 2-driven inflammation defines major subphenotypes of asthma. *Am. J. Respir. Crit. Care Med.* **180**, 388–395 (2009).
5. Orange, D. E. *et al.* Identification of Three Rheumatoid Arthritis Disease Subtypes by Machine Learning Integration of Synovial Histologic Features and RNA Sequencing Data. *Arthritis & rheumatology (Hoboken, N.J.)* **70**, (2018).
6. Zheng, C. & Xu, R. Molecular subtyping of Alzheimer's disease with consensus non-negative matrix factorization. *PLoS One* **16**, e0250278 (2021).
7. Zhang, F. *et al.* Deconstruction of rheumatoid arthritis synovium defines inflammatory subtypes. *Nature* **623**, 616–624 (2023).
8. McCarthy, M. I. Painting a new picture of personalised medicine for diabetes. *Diabetologia* **60**, 793–799 (2017).
9. Schork, N. J. Personalized medicine: Time for one-person trials. *Nature* **520**, (2015).
10. Seyhan, A. A. Lost in translation: the valley of death across preclinical and clinical divide – identification of problems and overcoming obstacles. *Transl. Med. Commun.* **4**, (2019).
11. Brunet, J.-P., Tamayo, P., Golub, T. R. & Mesirov, J. P. Metagenes and molecular pattern discovery using matrix factorization. *Proc. Natl. Acad. Sci. U. S. A.* **101**, 4164–4169 (2004).
12. Yang, Z. & Michailidis, G. A non-negative matrix factorization method for detecting modules in heterogeneous omics multi-modal data. *Bioinformatics* **32**, 1–8 (2016).
13. Zhang, S. *et al.* Discovery of multi-dimensional modules by integrative analysis of cancer genomic data. *Nucleic Acids Res.* **40**, 9379–9391 (2012).
14. Cancer Genome Atlas Research Network. Integrated genomic analyses of ovarian carcinoma. *Nature* **474**, 609–615 (2011).
15. Gao, Y. & Church, G. Improving molecular cancer class discovery through sparse

non-negative matrix factorization. *Bioinformatics* **21**, 3970–3975 (2005).

16. Pontes, B., Giráldez, R. & Aguilar-Ruiz, J. S. Biclustering on expression data: A review. *J. Biomed. Inform.* **57**, 163–180 (2015).
17. Padilha, V. A. & Campello, R. J. G. B. A systematic comparative evaluation of biclustering techniques. *BMC Bioinformatics* **18**, 55 (2017).
18. Xie, J., Ma, A., Fennell, A., Ma, Q. & Zhao, J. It is time to apply biclustering: a comprehensive review of biclustering applications in biological and biomedical data. *Brief. Bioinform.* **20**, 1449–1464 (2019).
19. Castanho, E. N., Aidos, H. & Madeira, S. C. Biclustering data analysis: a comprehensive survey. *Brief Bioinform* **25**, (2024).
20. López-Fernández, A., Gomez-Vela, F. A., Rodriguez-Baena, D. S., Delgado-Chaves, F. M. & Gonzalez-Dominguez, J. Biclustering in bioinformatics using big data and High Performance Computing applications: challenges and perspectives, a review. *J. Supercomput.* **81**, (2025).
21. Tini, G., Marchetti, L., Priami, C. & Scott-Boyer, M.-P. Multi-omics integration—a comparison of unsupervised clustering methodologies. *Brief. Bioinform.* **20**, 1269–1279 (2017).
22. Chauvel, C., Novoloaca, A., Veyre, P., Reynier, F. & Becker, J. Evaluation of integrative clustering methods for the analysis of multi-omics data. *Brief. Bioinform.* **21**, 541–552 (2020).
23. Duan, R. *et al.* CEPICS: A Comparison and Evaluation Platform for Integration Methods in Cancer Subtyping. *Front. Genet.* **10**, (2019).
24. Rappoport, N. & Shamir, R. Multi-omic and multi-view clustering algorithms: review and cancer benchmark. *Nucleic Acids Res.* **46**, 10546–10562 (2018).
25. Duan, R. *et al.* Evaluation and comparison of multi-omics data integration methods for cancer subtyping. *PLoS Comput. Biol.* **17**, e1009224 (2021).
26. Leng, D. *et al.* A benchmark study of deep learning-based multi-omics data fusion methods for cancer. *Genome Biol.* **23**, 1–32 (2022).

27. Zhang, T., Ramakrishnan, R. & Livny, M. BIRCH: an efficient data clustering method for very large databases. *SIGMOD Rec.* **25**, 103–114 (1996).

28. Ester, M., Kriegel, H., Sander, J. & Xu, X. A density-based algorithm for discovering clusters in large spatial databases with noise. *KDD* 226–231 (1996).

29. Lee, D. D. & Seung, H. S. Learning the parts of objects by non-negative matrix factorization. *Nature* **401**, 788–791 (1999).

30. Roussinov, D. G. & Chen, H. Document clustering for electronic meetings: an experimental comparison of two techniques. *Decis. Support Syst.* **27**, 67–79 (1999).

31. Shi, J. & Malik, J. Normalized cuts and image segmentation. *IEEE Trans. Pattern Anal. Mach. Intell.* **22**, 888–905 (2000).

32. Bar-Joseph, Z., Gifford, D. K. & Jaakkola, T. S. Fast optimal leaf ordering for hierarchical clustering. *Bioinformatics* **17 Suppl 1**, S22–9 (2001).

33. Comaniciu, D. & Meer, P. Mean shift: a robust approach toward feature space analysis. *IEEE Trans. Pattern Anal. Mach. Intell.* **24**, 603–619 (2002).

34. Frey, B. J. & Dueck, D. Clustering by passing messages between data points. *Science* **315**, 972–976 (2007).

35. Zou, H., Hastie, T. & Tibshirani, R. Sparse Principal Component Analysis. *J. Comput. Graph. Stat.* **15**, 265–286 (2006).

36. Meng, C., Helm, D., Frejno, M. & Kuster, B. moCluster: Identifying Joint Patterns Across Multiple Omics Data Sets. *J. Proteome Res.* **15**, 755–765 (2016).

37. Mo, Q. & Shen, R. iClusterPlus: Integrative clustering of multi-type genomic data. *Bioconductor R package version* **1**, (2018).

38. Argelaguet, R. *et al.* Multi-Omics Factor Analysis-a framework for unsupervised integration of multi-omics data sets. *Mol. Syst. Biol.* **14**, e8124 (2018).

39. Scrucca, L., Fraley, C., Brendan Murphy, T. & Raftery, A. E. *Model-Based Clustering, Classification, and Density Estimation Using Mclust in R*. (CRC Press, 2023).

40. Lazzeroni, L. & Owen, A. PLAID MODELS FOR GENE EXPRESSION DATA. *Stat. Sin.* **12**, 61–86 (2002).

41. Prelić, A. *et al.* A systematic comparison and evaluation of biclustering methods for gene expression data. *Bioinformatics* **22**, 1122–1129 (2006).
42. Huttenhower, C. *et al.* Detailing regulatory networks through large scale data integration. *Bioinformatics* **25**, 3267–3274 (2009).
43. Li, G., Ma, Q., Tang, H., Paterson, A. H. & Xu, Y. QUBIC: a qualitative biclustering algorithm for analyses of gene expression data. *Nucleic Acids Res.* **37**, e101 (2009).
44. Hochreiter, S. *et al.* FABIA: factor analysis for bicluster acquisition. *Bioinformatics* **26**, 1520–1527 (2010).
45. Csárdi, G., Kutalik, Z. & Bergmann, S. Modular analysis of gene expression data with R. *Bioinformatics* **26**, 1376–1377 (2010).
46. Serin, A. & Vingron, M. DeBi: Discovering Differentially Expressed Biclusters using a Frequent Itemset Approach. *Algorithms Mol. Biol.* **6**, 1–12 (2011).
47. Langfelder, P. & Horvath, S. WGCNA: an R package for weighted correlation network analysis. *BMC Bioinformatics* **9**, 559 (2008).
48. Zhang, B. & Horvath, S. A general framework for weighted gene co-expression network analysis. *Stat. Appl. Genet. Mol. Biol.* **4**, Article17 (2005).
49. Blondel, V. D., Guillaume, J.-L., Lambiotte, R. & Lefebvre, E. Fast unfolding of communities in large networks. *arXiv [physics.soc-ph]* (2008) doi:10.48550/ARXIV.0803.0476.
50. Traag, V. A., Waltman, L. & van Eck, N. J. From Louvain to Leiden: guaranteeing well-connected communities. *Sci Rep* **9**, 5233 (2019).
51. Newman, M. E. J. & Girvan, M. Finding and evaluating community structure in networks. *Phys. Rev. E Stat. Nonlin. Soft Matter Phys.* **69**, 026113 (2004).
52. Ritchie, M. E. *et al.* limma powers differential expression analyses for RNA-sequencing and microarray studies. *Nucleic Acids Res.* **43**, e47 (2015).
53. Zolotareva, O. *et al.* Identification of differentially expressed gene modules in heterogeneous diseases. *Bioinformatics* **37**, 1691–1698 (2021).
54. Singhania, A. *et al.* Multitissue Transcriptomics Delineates the Diversity of Airway T Cell

Functions in Asthma. *Am. J. Respir. Cell Mol. Biol.* **58**, 261–270 (2018).

55. Perou, C. M. *et al.* Molecular portraits of human breast tumours. *Nature* **406**, 747–752 (2000).
56. Parker, J. S. *et al.* Supervised risk predictor of breast cancer based on intrinsic subtypes. *J. Clin. Oncol.* **27**, 1160–1167 (2009).
57. Comprehensive molecular portraits of human breast tumours. *Nature* **490**, 61–70 (2012).
58. Curtis, C. *et al.* The genomic and transcriptomic architecture of 2,000 breast tumours reveals novel subgroups. *Nature* **486**, 346–352 (2012).
59. Wenzel, S. E. Asthma phenotypes: the evolution from clinical to molecular approaches. *Nat. Med.* **18**, 716–725 (2012).
60. Modena, B. D. *et al.* Gene Expression Correlated with Severe Asthma Characteristics Reveals Heterogeneous Mechanisms of Severe Disease. *Am. J. Respir. Crit. Care Med.* **195**, 1449–1463 (2017).
61. Spira, A. *et al.* Effects of cigarette smoke on the human airway epithelial cell transcriptome. *Proc. Natl. Acad. Sci. U. S. A.* **101**, 10143–10148 (2004).
62. Wu, Q. *et al.* Cancer-associated adipocytes: key players in breast cancer progression. *J Hematol Oncol* **12**, 95 (2019).
63. Chin, K. *et al.* Genomic and transcriptional aberrations linked to breast cancer pathophysiologies. *Cancer Cell* **10**, 529–541 (2006).
64. Todorović-Raković, N. & Whitfield, J. R. Therapeutic implications of the interplay between interferons and ER in breast cancer. *Cytokine Growth Factor Rev.* **75**, 119–125 (2024).
65. Naik, A., Lattab, B., Qasem, H. & Decock, J. Cancer testis antigens: Emerging therapeutic targets leveraging genomic instability in cancer. *Mol. Ther. Oncol.* **32**, 200768 (2024).
66. Trevisi, E. *et al.* Neuroendocrine breast carcinoma: a rare but challenging entity. *Med. Oncol.* **37**, 70 (2020).

67. Gezelius, E. *et al.* Seizure-related homolog protein 6 (SEZ6): Biology and therapeutic target in neuroendocrine carcinomas. *Clin. Cancer Res.* **31**, 4419–4428 (2025).
68. Sato, Y. *et al.* Integrated molecular analysis of clear-cell renal cell carcinoma. *Nat. Genet.* **45**, 860–867 (2013).
69. Brannon, A. R. *et al.* Molecular Stratification of Clear Cell Renal Cell Carcinoma by Consensus Clustering Reveals Distinct Subtypes and Survival Patterns. *Genes Cancer* **1**, 152–163 (2010).
70. Jiang, Z. *et al.* Analysis of RNA-binding protein IMP3 to predict metastasis and prognosis of renal-cell carcinoma: a retrospective study. *Lancet Oncol* **7**, 556–564 (2006).
71. Hoffmann, N. E. *et al.* External validation of IMP3 expression as an independent prognostic marker for metastatic progression and death for patients with clear cell renal cell carcinoma. *Cancer* **112**, 1471–1479 (2008).
72. Zhang, Q. *et al.* Troponin T1 in tumorigenesis and immune modulation: Insights into multiple cancers and kidney renal clear cell carcinoma. *J Cell Mol Med* **28**, e18410 (2024).
73. Wang, S. *et al.* PTPRH promotes the progression of non-small cell lung cancer via glycolysis mediated by the PI3K/AKT/mTOR signaling pathway. *J Transl Med* **21**, 819 (2023).
74. Jiang, H., Chen, H., Wan, P., Song, S. & Chen, N. Downregulation of enhancer RNA EMX2OS is associated with poor prognosis in kidney renal clear cell carcinoma. *Aging (Albany NY)* **12**, 25865–25877 (2020).
75. Zhou, X. *et al.* EMX2 inhibits clear cell renal cell carcinoma progress via modulating Akt/FOXO3a pathway. *Mol Carcinog* **63**, 951–961 (2024).
76. Pelz, L. *et al.* Semi-continuous Propagation of Influenza A Virus and Its Defective Interfering Particles: Analyzing the Dynamic Competition To Select Candidates for Antiviral Therapy. *J. Virol.* **95**, e0117421 (2021).
77. Skinnider, M. A. *et al.* Proteomic Portraits Reveal Evolutionarily Conserved and

Divergent Responses to Spinal Cord Injury. *Mol Cell Proteomics* **20**, 100096 (2021).

78. Goldman, M. J. *et al.* Visualizing and interpreting cancer genomics data via the Xena platform. *Nat. Biotechnol.* **38**, 675–678 (2020).
79. Wolf, F. A., Angerer, P. & Theis, F. J. SCANPY: large-scale single-cell gene expression data analysis. *Genome Biol* **19**, 15 (2018).
80. Engesser, J. *et al.* Immune profiling-based targeting of pathogenic T cells with ustekinumab in ANCA-associated glomerulonephritis. *Nat Commun* **15**, 8220 (2024).
81. Cancer Genome Atlas Research Network. Comprehensive molecular characterization of clear cell renal cell carcinoma. *Nature* **499**, 43–49 (2013).
82. Liang, W.-W. *et al.* Integrative multi-omic cancer profiling reveals DNA methylation patterns associated with therapeutic vulnerability and cell-of-origin. *Cancer Cell* **41**, 1567–1585.e7 (2023).
83. Satija, R., Farrell, J. A., Gennert, D., Schier, A. F. & Regev, A. Spatial reconstruction of single-cell gene expression data. *Nat Biotechnol* **33**, 495–502 (2015).
84. Cerami, E. *et al.* The cBio cancer genomics portal: an open platform for exploring multidimensional cancer genomics data. *Cancer Discov.* **2**, 401–404 (2012).
85. Gao, J. *et al.* Integrative analysis of complex cancer genomics and clinical profiles using the cBioPortal. *Sci. Signal.* **6**, l1 (2013).
86. Woodruff, P. G. *et al.* Genome-wide profiling identifies epithelial cell genes associated with asthma and with treatment response to corticosteroids. *Proc. Natl. Acad. Sci. U. S. A.* **104**, (2007).
87. Barrett, T. *et al.* NCBI GEO: archive for functional genomics data sets--update. *Nucleic Acids Res.* **41**, (2013).
88. Davis, S. & Meltzer, P. S. GEOquery: a bridge between the Gene Expression Omnibus (GEO) and BioConductor. *Bioinformatics* **23**, 1846–1847 (2007).
89. Gautier, L., Cope, L., Bolstad, B. M. & Irizarry, R. A. affy—analysis of Affymetrix GeneChip data at the probe level. *Bioinformatics* **20**, 307–315 (2004).
90. Durinck, S., Spellman, P. T., Birney, E. & Huber, W. Mapping identifiers for the

integration of genomic datasets with the R/Bioconductor package biomaRt. *Nat. Protoc.* **4**, 1184–1191 (2009).

91. Fougner, C., Bergholtz, H., Norum, J. H. & Sørlie, T. Re-definition of claudin-low as a breast cancer phenotype. *Nat. Commun.* **11**, 1787 (2020).

92. Rose, T. D. *et al.* MoSBi: Automated signature mining for molecular stratification and subtyping. *Proc. Natl. Acad. Sci. U. S. A.* **119**, e2118210119 (2022).

93. Rocha, O. & Mendes, R. JBiclustGE: Java API with unified biclustering algorithms for gene expression data analysis. *Knowl. Based Syst.* **155**, 83–87 (2018).

94. Pandey, G., Atluri, G., Steinbach, M., Myers, C. L. & Kumar, V. An association analysis approach to biclustering. in *Proceedings of the 15th ACM SIGKDD international conference on Knowledge discovery and data mining* 677–686 (Association for Computing Machinery, New York, NY, USA, 2009).

95. Davidson-Pilon, C. lifelines: survival analysis in Python. *J. Open Source Softw.* **4**, 1317 (2019).

96. Robinson, M. D., McCarthy, D. J. & Smyth, G. K. edgeR: a Bioconductor package for differential expression analysis of digital gene expression data. *Bioinformatics* **26**, 139–140 (2010).

97. Yu, G., Wang, L.-G., Han, Y. & He, Q.-Y. clusterProfiler: an R package for comparing biological themes among gene clusters. *OMICS* **16**, 284–287 (2012).

98. Ashburner, M. *et al.* Gene ontology: tool for the unification of biology. The Gene Ontology Consortium. *Nat. Genet.* **25**, 25–29 (2000).

99. Gene Ontology Consortium *et al.* The Gene Ontology knowledgebase in 2023. *Genetics* **224**, (2023).

100. Schriml, L. M. *et al.* Human Disease Ontology 2018 update: classification, content and workflow expansion. *Nucleic Acids Res.* **47**, D955–D962 (2019).

101. Kanehisa, M., Furumichi, M., Sato, Y., Kawashima, M. & Ishiguro-Watanabe, M. KEGG for taxonomy-based analysis of pathways and genomes. *Nucleic Acids Res.* **51**, D587–D592 (2023).

102. Milacic, M. *et al.* The Reactome Pathway Knowledgebase 2024. *Nucleic Acids Res.* **52**, D672–D678 (2024).

103. Bullard, J. H., Purdom, E., Hansen, K. D. & Dudoit, S. Evaluation of statistical methods for normalization and differential expression in mRNA-Seq experiments. *BMC Bioinformatics* **11**, 1–13 (2010).

104. Benjamini, Y. & Hochberg, Y. Controlling the false discovery rate: A practical and powerful approach to multiple testing. *J. R. Stat. Soc. Series B Stat. Methodol.* **57**, 289–300 (1995).